\documentclass[fonts]{icst}

\usepackage{moreverb}

\usepackage[breaklinks,colorlinks,bookmarksopen,bookmarksnumbered,linkcolor=ICSTblue,citecolor=blue,urlcolor=ICSTblue]{hyperref}
\usepackage{breakurl}
\usepackage{doi}
\usepackage{graphicx}
\usepackage{amssymb,amsmath,amsfonts}
\usepackage{algorithm,algpseudocode}
\usepackage{verbatim}
\usepackage{url}
\usepackage{multirow}
\usepackage{subfig}

\newcommand{\taba}{
\begin{tabular}{cc|c|c|l}
\cline{3-4}
& & \multicolumn{2}{ c| }{Predicted} \\ \cline{3-4}
& & NO & YES \\ \cline{1-4}
\multicolumn{1}{ |c  }{\multirow{2}{*}{Actual} } &
\multicolumn{1}{ |c| }{NO}  &$17981$ &$23129$ &     \\ \cline{2-4}
\multicolumn{1}{ |c  }{}                        &
\multicolumn{1}{ |c| }{YES}  &$57915$& $116583$ &     \\ \cline{1-4}
\end{tabular}
}

\newcommand{\tabb}{
\begin{tabular}{cc|c|c|l}
\cline{3-4}
& & \multicolumn{2}{ c| }{Predicted} \\ \cline{3-4}
& & NO & YES \\ \cline{1-4}
\multicolumn{1}{ |c  }{\multirow{2}{*}{Actual} } &
\multicolumn{1}{ |c| }{NO}  &$3437$ &$37673$ &     \\ \cline{2-4}
\multicolumn{1}{ |c  }{}                        &
\multicolumn{1}{ |c| }{YES}  &$9810$& $164688$ &     \\ \cline{1-4}
\end{tabular}
}

\newcommand{\tabc}{
\begin{tabular}{cc|c|c|l}
\cline{3-4}
& & \multicolumn{2}{ c| }{Predicted} \\ \cline{3-4}
& & NO & YES \\ \cline{1-4}
\multicolumn{1}{ |c  }{\multirow{2}{*}{Actual} } &
\multicolumn{1}{ |c| }{NO}  &$5191$ &$35919$ &     \\ \cline{2-4}
\multicolumn{1}{ |c  }{}                        &
\multicolumn{1}{ |c| }{YES}  &$11394$& $163104$ &     \\ \cline{1-4}
\end{tabular}
}

\newcommand{\tabd}{
\begin{tabular}{cc|c|c|l}
\cline{3-4}
& & \multicolumn{2}{ c| }{Predicted} \\ \cline{3-4}
& & NO & YES \\ \cline{1-4}
\multicolumn{1}{ |c  }{\multirow{2}{*}{Actual} } &
\multicolumn{1}{ |c| }{NO}  &$4176$ &$36934$ &     \\ \cline{2-4}
\multicolumn{1}{ |c  }{}                        &
\multicolumn{1}{ |c| }{YES}  &$13173$& $161325$ &     \\ \cline{1-4}
\end{tabular}
}

\newcommand{\tabe}{
\begin{tabular}{cc|c|c|l}
\cline{3-4}
& & \multicolumn{2}{ c| }{Predicted} \\ \cline{3-4}
& & NO & YES \\ \cline{1-4}
\multicolumn{1}{ |c  }{\multirow{2}{*}{Actual} } &
\multicolumn{1}{ |c| }{NO}  &$457$ &$38653$ &     \\ \cline{2-4}
\multicolumn{1}{ |c  }{}                        &
\multicolumn{1}{ |c| }{YES}  &$532$& $175966$ &     \\ \cline{1-4}
\end{tabular}
}

\newcommand\BibTeX{{\rmfamily B\kern-.05em \textsc{i\kern-.025em b}\kern-.08em
T\kern-.1667em\lower.7ex\hbox{E}\kern-.125emX}}

\def\volumeyear{2014}

\journalname{XXXXXX}
\articletype{Research Article \textbf{EAI.EU}}
 \def\volumeyear{XXXX}
 \def\volumenumber{XX}
  \def\issuemonth{XX-XX}
  \def\issuenumber{X}
  \def\articleid{XX}
  \def\copyrightholder{Patgiri\textit{ et al.}, licensed to ICST}
  \copyrightnote{This is an open access article distributed under the terms of the Creative Commons Attribution license (\url{http://creativecommons.org/licenses/by/3.0/}), which permits unlimited use, distribution and reproduction in any medium so long as the original work is properly cited.}
  \def\articledoi{10.4108/XX.X.X.XX}
\received{XXXX}
  \accepted{XXXX}
  \published{XXXX}

\begin{document}

\runningheads{Patgiri et al.}{Empirical Study on Airline Delay Analysis and Prediction}

\title{Empirical Study on Airline Delay Analysis and Prediction}

\author{Ripon Patgiri\affil{1}\fnoteref{1}, Sajid Hussain\affil{1}\fnoteref{1}, and Aditya Nongmeikapam\affil{1}\fnoteref{1}}

\address{\affilnum{1} National Institute of Technology Silchar, Assam-788010, India}

\abstract{The Big Data analytics are a logical analysis of very large scale datasets. The data analysis enhances an organization and improve the decision making process. In this article, we present Airline Delay Analysis and Prediction to analyze airline datasets with the combination of weather dataset. In this research work, we consider various attributes to analyze flight delay, for example, day-wise, airline-wise, cloud cover, temperature, etc. Moreover, we present rigorous experiments on various machine learning model to predict correctly the delay of a flight, namely, logistic regression with L2 regularization, Gaussian Naive Bayes, K-Nearest Neighbors, Decision Tree classifier and Random forest model.  The accuracy of the Random Forest model is 82\% with a delay threshold of 15 minutes of flight delay. The analysis is carried out using dataset from 1987 to 2008, the training is conducted with dataset from 2000 to 2007 and validated prediction result using 2008 data. Moreover, we have got recall 99\% in the Random Forest model.}



\fnotetext[1]{Ripon Patgiri, Corresponding author.  Email: \email{ripon@cse.nits.ac.in, sajidhussain1995@gmail.com, adityank.adi@gmail.com}}

\maketitle

\section{Introduction}
MapReduce \cite{MR04} is the most famous programming engine for developing parallel and distributed application. It is the ideal programming language for large-scale computation with low-cost commodity hardware. It is serving the data-intensive computing since its inception. MapReduce consists of two functions, namely, map and reduce; yet it is a very powerful programing engine. Moreover, MapReduce is also strictly key/value pair programming model. Interestingly, every problem in data-intensive computation can be converted into key/value pair easily. MapReduce boomed in the IT industry and change the game of the industries. It is applied in many fields in Computer Science and it is also applied in interdisciplinary computing nowadays. For instance, the genome project. MapReduce has left nothing untouched for large-scale computation. The MapReduce programming model is easier to develop more customized framework using map and reduce function. Thus, airline delay analysis is performed using MapReduce, which takes few seconds to complete. Besides, other Big Data tools can also be used, for instance, Apache Spark \cite{Zaharia}.

The airline data is growing day-by-day along with the number of new airlines, due to the ever growing population of air travelers. There is always a quest for airline data analytics to support more realistic data analysis with the combination of weather data \cite{Choi}. The MapReduce programming model is used to experiment the behavior of flight delay. In addition, analysis is performed using the dataset obtained from American Statistical Association \cite{ASA} and National Weather Service’s National Digital Forecast Database (NDFD) \cite{NDFD}. The airline dataset and weather dataset from 1987 to 2008 is huge which cannot be processed in conventional system. Therefore, MapReduce is the best programming model to develop the airline delay analytics. Moreover, parallel processing with low-cost commodity hardware is easier to experiment than HPC, and better than a conventional systems. 

The prediction is performed using various Data Mining techniques which are used to uncover hidden truth \cite{Sharma}. We deploys conventional Machine learning models, and the comparison of the results as shown in table ~\ref{tab}, namely, Logistic regression \cite{Cox} with L2 regularization, K-Nearest Neighbor \cite{Cover}, Gaussian Naive Bayes \cite{John}, decision tree classifier \cite{Quinlan}, and Random Forest \cite{Ho,RForest}. The random forest model outperforms the other four models. However, prediction is a very time-consuming process. As per our experience, more training data makes the prediction more accurate. However, we reduce the training data and just used the dataset from 2000 to 2007 to examine the accuracy of the Machine Learning algorithms. The prediction is evaluated using the data of 2008.

\begin{table*}[!ht]
\centering
\caption{Comparison of Machine Learning models}
\label{tab}
\begin{tabular}{|p{4cm}|c|c|c|c|}
\hline
\textbf{Name} & \textbf{Precision} & \textbf{Recall}  & \textbf{F1 score} & \textbf{Accuracy} \\ \hline
Logistic Regression with L2 regularization & 0.83 & 0.67&0.74 & 0.62 \\ \hline
K-Nearest Neighbors & 0.81 & 0.94 & 0.87 & 0.78\\ \hline
Gaussian Naive Bayes & 0.82 & 0.93 & 0.87 & 0.78\\ \hline
Decision Tree Classifier & 0.81 & 0.92 & 0.87 & 0.77 \\ \hline
Random Forest & 0.82 & 0.99 & 0.90 & 0.82 \\ \hline
\end{tabular}
\end{table*}

\subsection{Contribution}
The contribution of this work is summarized as follows-
\begin{itemize}
\item Development of Airline Analytics, which analyze the large dataset for flight delays. 
\item Merging of the weather dataset with the airline dataset to get more realistic results.
\item Analysis report generation helps in the decision making process.
\item Prediction of flight delay using various machine learning models.
\end{itemize}

\subsection{Organization}
The paper is organized as follows- section ~\ref{rw} discusses on the background and related work on flight delay analysis and prediction. Section ~\ref{ds} exposes the data sources and various attributes of the dataset in analysis and prediction. Section ~\ref{da} provides detailed analysis of airline delay with weather data. Section ~\ref{pr} predicts the delay of flights using various available machine learning models. Finally, section ~\ref{con} concludes the work.

\section{Background and Related work}
\label{rw}
The delay in airlines is very frequent, and it occurs due to weather conditions. The airline delay analysis is an important analysis to know the behavior of a flight and its future probable delay. There are numerous opportunities to work for the airline industry. The realistic airline data are available for free to experiment and develop interesting solutions to existing dilemmas. However, similar work has been done in the article \cite{Rebo,Fergu,Choi,Bloem,Bel} for delay analysis. The article \cite{Liu} analyze the latency of air traffic. There are various parameters to consider in analyzing the air traffic data. More precisely, the main motivation is to analyze the airline delay using the combination of weather data and air traffic data. This creates large sized air traffic data to analyze. Rebollo and Balakrishnan \cite{Rebo} provides 100 most-delayed origin-destination pairs from operation dataset.  Bloem et al. \cite{Bloem} works on various cost functions of the airline and analyses the delay. Ferguson et al. \cite{Fergu} estimate the cost of a delay of flight. Choi et al. \cite{Choi} merges weather data and airline data. Choi et al. \cite{Choi} achieves 80.36 \% accuracy of the Random Forest model using 2005 to 2015 airline and weather data. Belcastro et al. \cite{Bel} designs a predictor based on MapReduce. Belcastro et al. \cite{Bel} achieves 74.2\% accuracy with the delay threshold of 15 minutes, however,  the Random Forest model outperforms this accuracy.

\subsection{Logistic Regression}
The Logistic Regression technique is developed by D. R. Cox \cite{Cox} in 1958. After its invention, logistic regression become an integral part of Machine Learning. The purest form of Logistic Regression by itself is very much susceptible to the problem of overfitting. To get rid of this problem, Logistic Regression is often used with L2 regularization. This leads to more precise and accurate predictions, with the overfitting taken care of.

\subsection{Gaussian Naive Bayes}
It's a Machine Learning Technique that uses a probabilistic approach for classification of records. It is one of the two types of Gaussian Classifiers \cite{John}. It assumes Conditional Independence of the different attributes of the data used for the classification. The second type of Gaussian Classifier is called the BBN (Bayesian Belief Network), it performs the classification task without the help of the assumption of Conditional Independence. In the Naive Bayes Classifier, we find out the probability of occurrence of each of the possible classes given the values of the particular attributes. Then, the class for which this probability would turn out to be the largest will be declared as the target class for the given attribute values.

For example, let us consider a dataset with two possible target classes, namely 'yes' and 'no' and two attributes 'A' and 'B'. And, the attribute values given are $A = a$ and $B = b$. So, the probabilities to be calculated are :

\[ P(yes/A = a, B = b) \]
and 
\[P(no/A = a, B = b)\]
Now, 
\[P(yes/A = a, B = b)\]
\[= P(A = a, B = b/yes)P(yes)/P(A = a, B = b)\]
\[= P(A = a/yes)P(B = b/yes)P(yes)/k\]

Here, we have considered the marginal probability \[P(A = a, B = b)\] to be a constant k, since it will be same for all the classes.

\subsection{K-nearest neighbors}
The K-Nearest Neighbors (KNN) is a classifier which clusters K nearest items \cite{Cover}. This KNN technique is extremely useful in learning and prediction. The total nearest neighbors are taken as 10 which provides us the best results. It locates the nearest neighbors in the data point space, and then, labels the unknown data point as the same class label, as that of the known class label.

\subsection{Decision Tree}
The decision tree is one of the most basic Machine Learning Techniques \cite{Quinlan}. The key feature of the decision tree is the impurities in the dataset. These impurities suggest how heterogeneous or homogeneous a given set of data is. If the impurities turn out to be zero, it means that the data set is classified. These impurities are found by calculating entropy, gini index, misclassification error, etc for the given data.

\begin{figure}[ht]
\centering
\includegraphics[width=0.45\textwidth]{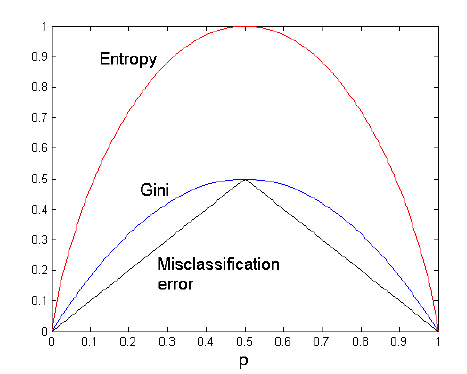}
\caption{Relation among Entropy, Gini Index and Misclassification error.}
\end{figure}

Entropy- \[ E(t) = -\sum_i(p(i/t)\log_2(p(i/t)))\]

Gini Index-\[ Gini(t) = 1 - \sum_i(p(i/t))^2\]

Misclassification error-\[error(t) = 1 - \max_i(p(i/t))\]

Where, $p(i/t)$ is the fraction of records belonging to class $i$ at node $t$ and $i$ varies from $0$ to $c-1$. $c$ is the total number of possible classes.

\subsection{Random Forest}
The Random Forest creates a huge set of random trees which is very time consuming, but more accurate than other models in predicting \cite{Ho,RForest}. The Random Forest model is a set of decision tree created within a tree for better results. The features are selected randomly to form a sub-tree. Therefore, there are many random trees in training periods. These trees are bagged repeatedly for $T$ times. The Random Forest probability output is
\[p(c/v)=\frac{1}{T}\sum_t^Tp_t(c/v)\]
The total number of trees taken in the experiment is 50.

\subsection{Confusion Matrix}

\begin{table}[ht]
\centering
\caption{Confusion Matrix}
\begin{tabular}{cc|c|c|l}
\cline{3-4}
& & \multicolumn{2}{ c| }{Predicted} \\ \cline{3-4}
& & NO & YES \\ \cline{1-4}
\multicolumn{1}{ |c  }{\multirow{2}{*}{Actual} } &
\multicolumn{1}{ |c| }{NO}  &$\sum TN$ &$\sum FP$ &     \\ \cline{2-4}
\multicolumn{1}{ |c  }{}                        &
\multicolumn{1}{ |c| }{YES}  &$\sum FN $& $\sum TP$ &     \\ \cline{1-4}
\end{tabular}
\label{cm}
\end{table}

Table ~\ref{cm} shows the confusion matrix. The confusion matrix is composed of the number of actual delayed and predicted delayed flights \cite{Sharma1}. The True Negative (TN) is actually not delayed and predicted to be not delayed. The False Positive (FP) is actually not delayed and predicted to be delayed. The False Negative (FN) is actually delayed, but predicted to be not delayed. The True Positive (TP) is actually delayed and predicted to be delayed. The summation $\sum$ represents the total number of such events. Based on the Table ~\ref{cm}, precision, recall, F1 score and accuracy are calculated as follows-

\[Precision=\frac{\sum TP}{\sum FP+ \sum TP}\]
\[Recall=\frac{\sum TP}{\sum FN+\sum TP}\]
\[F1 score=2\times\frac{Precision\times Recall}{Precision+Recall}\]
\[Accuracy=\frac{\sum TN+\sum TP}{\sum TN+ \sum FP+ \sum FN+\sum TP}\]

\section{Data Source}
\label{ds}
The data source is the American Statistical Association \cite{ASA} and National Weather Service’s National Digital Forecast Database (NDFD) \cite{NDFD}. The data source contains US flight and weather data. We have taken data from 1987 to 2008 for analysis purpose and from 2000 to 2007 to train the models. Later, the resultant prediction is validated using 2008 dataset. There are various attributes in the joined dataset which are enlisted below-

\begin{itemize}
\item Year: 1987 to 2008
\item Month: 1 (January) to 12 (December)
\item Day of month: 1 to 31 
\item Day of week: 1 (Monday) to 7 (Sunday)
\item Unique carrier: Code of the airliner, e.g., AI (Air India)
\item Departure delay: The departure delay is measured in minutes.
\item Origin: The origin city of flight to the destination.
\item Destination: The destination city to reach by flight from origin.
\item Distance: The distance is measured in miles.
\item Maximum and minimum temperature: The degree of temperature is represented in Fahrenheit.
\item Maximum and minimum visibility: The visibility is represented in miles.
\item Mean wind speed: The mean wind speed is represented in m/s.
\item Precipitation: It is represented in inches.
\item Cloud cover: The cloud cover is fullness of cloud in the sky.
\end{itemize}

\section{Delay Analysis}
\label{da}

This work uses two different types of datasets, one data set gives the details regarding the delays in the airline industry and the other dataset gives the weather information for the different cities in the United States. This work is a combination of analysis of the delays encountered by the different airlines in services and the prediction of this delay for the future flights. The datasets are used for the analysis purpose. The predictor also uses the same datasets in filtered form. The first dataset is obtained from the American Statistical Association \cite{ASA} and it has the attributes are, namely, year, day of week, day of month, unique carrier name, distance traveled, scheduled and actual departure time, etc. The second dataset is obtained from the National Weather Service’s National Digital Forecast Database (NDFD) \cite{NDFD}. It contains information regarding the maximum and minimum temperature, visibility, wind speed, precipitation, and cloud cover. These two datasets are joined on the basis of the date using a MapReduce code and served to the prediction algorithms. The joined data are used for analysis purpose. Now, let’s discuss the two parts in details next.

\begin{table}[ht]
\centering
\caption{Experimental environment setup}
\begin{tabular}{|c|c|}
\hline
Name & Configuration \\ \hline
CPU &  i7-4770 CPU @ 3.40GHz\\ \hline
RAM &  4GB\\ \hline
HDD & 500 GB each \\ \hline
OS & Ubuntu 16.10 64-bit\\ \hline
Hadoop & V2.7.3 \\ \hline
Nodes & 20 \\ \hline
\end{tabular}
\label{ee}
\end{table}

The experimental environment is set up using the above configuration as depicted in Table ~\ref{ee}. Hadoop cluster is set up to conduct the experiment. The total number of nodes is 20 with 4GB RAM of each. This cluster is sufficient to join the dataset of airline and weather.

\begin{figure}[!ht]
    \centering
    \includegraphics[width=0.45\textwidth]{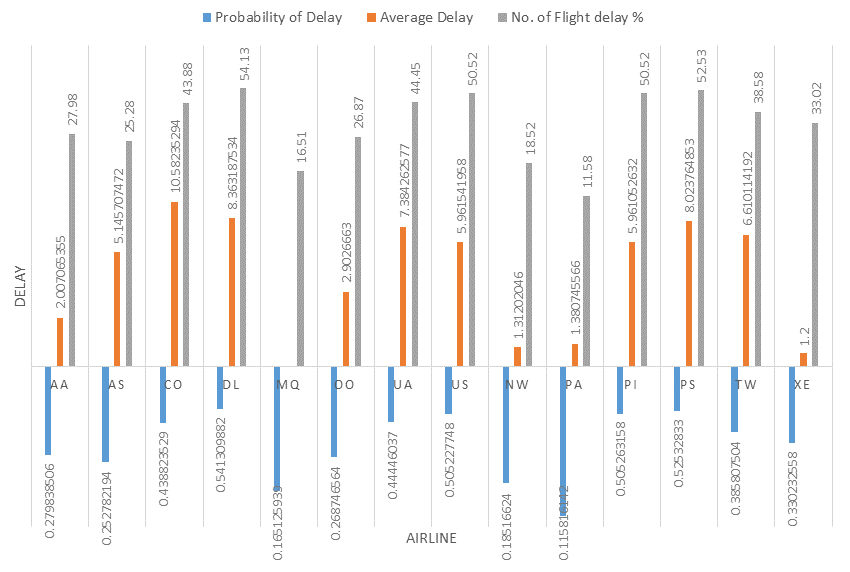}
    \caption{Airline-wise delay analysis without weather data. X-axis represents airlines and Y-axis represents number of delays.}
    \label{1}
\end{figure}

\begin{figure}[!ht]
\centering
\includegraphics[width=0.45\textwidth]{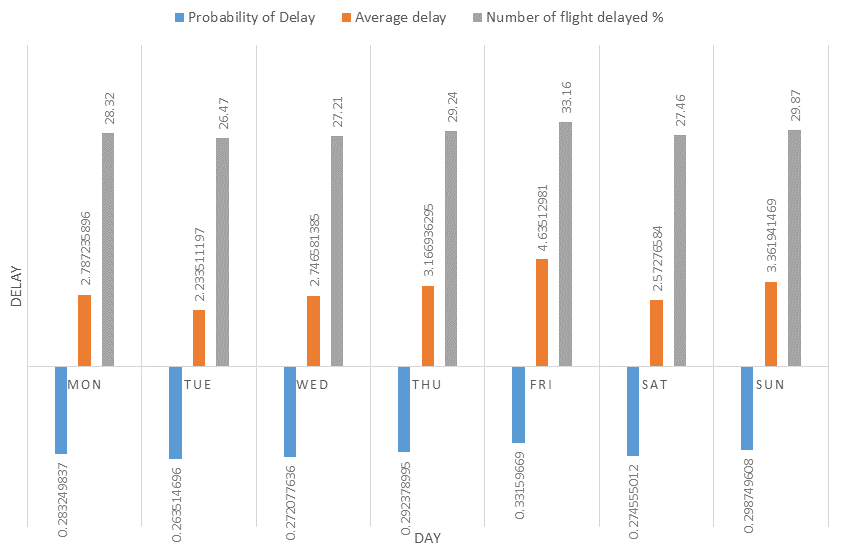}
\caption{Day-wise delay analysis without weather data. X-axis represents days and Y-axis represents number of delays.}	
\label{2}
\end{figure}

\begin{figure}[!ht]
\centering
\includegraphics[width=0.45\textwidth]{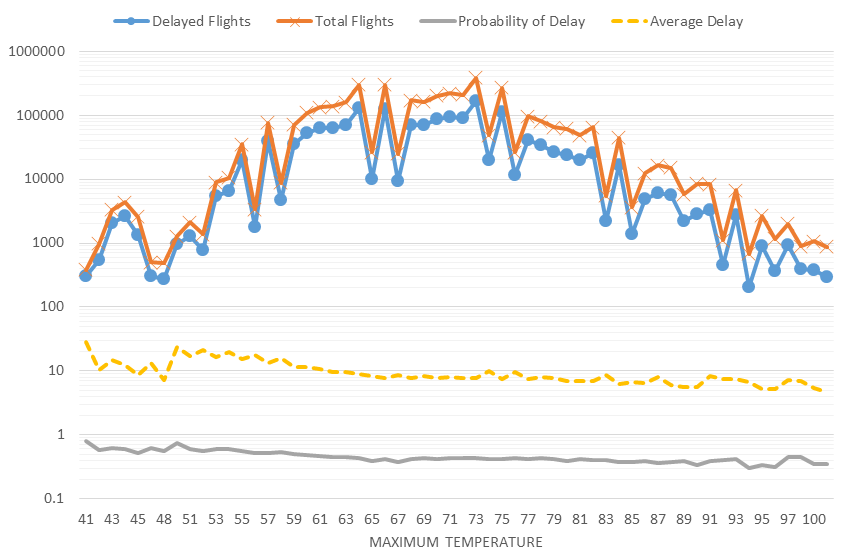}
\caption{Delay analysis on maximum temperature in Fahrenheit. X-axis represents maximum temperatures and Y-axis represents number of delays.}	
\label{3}
\end{figure}

\begin{figure}[!ht]
\centering
\includegraphics[width=0.45\textwidth]{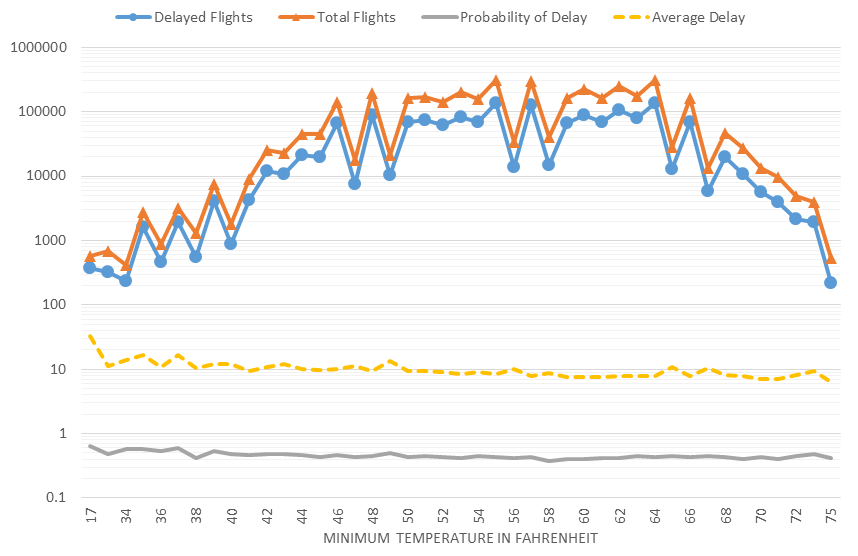}
\caption{Delay analysis on minimum temperature in Fahrenheit. X-axis represents minimum temperatures and Y-axis represents number of delays.}	
\label{4}
\end{figure}

\begin{figure}[!ht]
\centering
\includegraphics[width=0.45\textwidth]{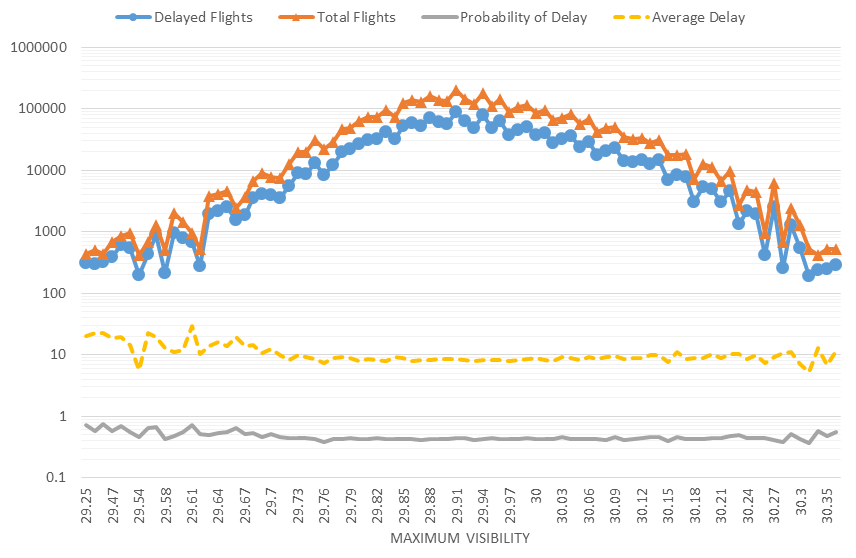}
\caption{Delay analysis on maximum visibility. X-axis represents maximum visibility and Y-axis represents number of delays.}
\label{5}
\end{figure}

\begin{figure}[!ht]
\centering
\includegraphics[width=0.45\textwidth]{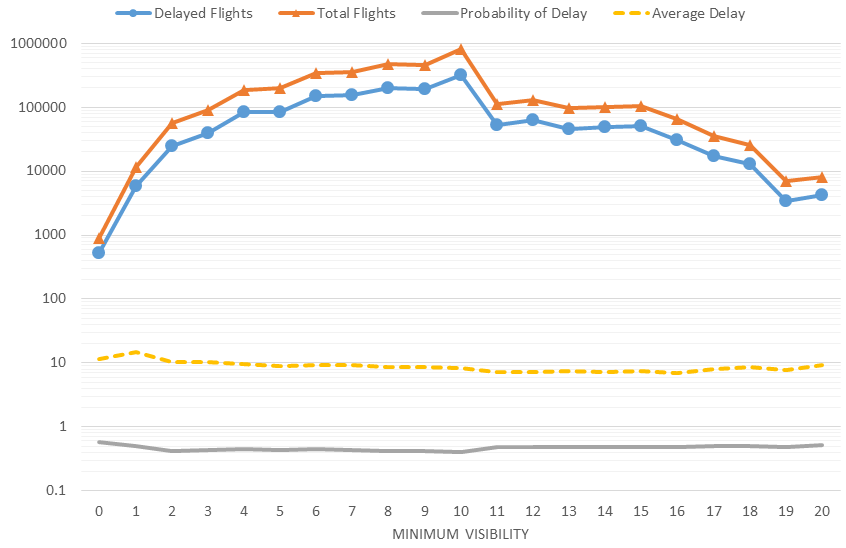}
\caption{Delay analysis on minimum visibility. X-axis represents minimum visibility and Y-axis represents number of delays.}
\label{6}
\end{figure}

\begin{figure}[!ht]
\centering
\includegraphics[width=0.45\textwidth]{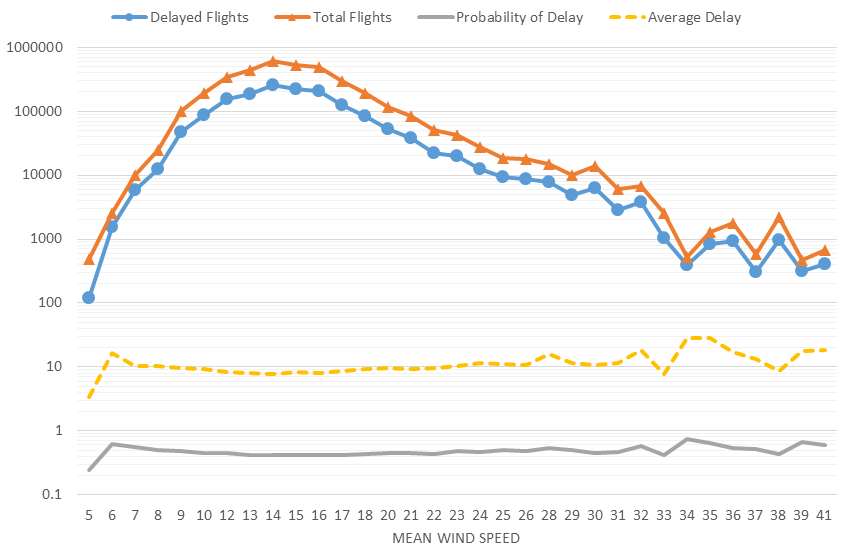}
\caption{Delay analysis on mean wind speed. X-axis represents mean wind speed and Y-axis represents number of delays.}
\label{7}
\end{figure}

\begin{figure}[!ht]
\centering
\includegraphics[width=0.45\textwidth]{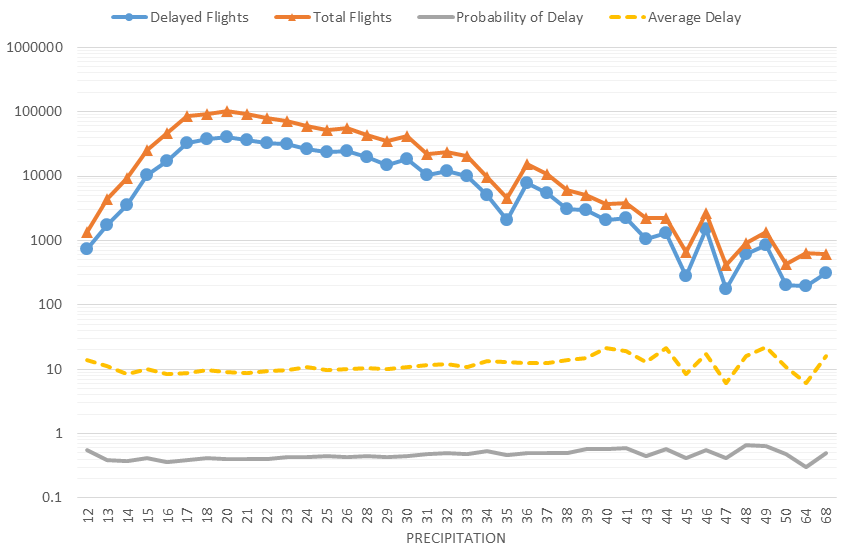}
\caption{Delay analysis on precipitation. X-axis represents precipitation and Y-axis represents number of delays.}
\label{8}
\end{figure}
	
\begin{figure}[!ht]
\centering
\includegraphics[width=0.45\textwidth]{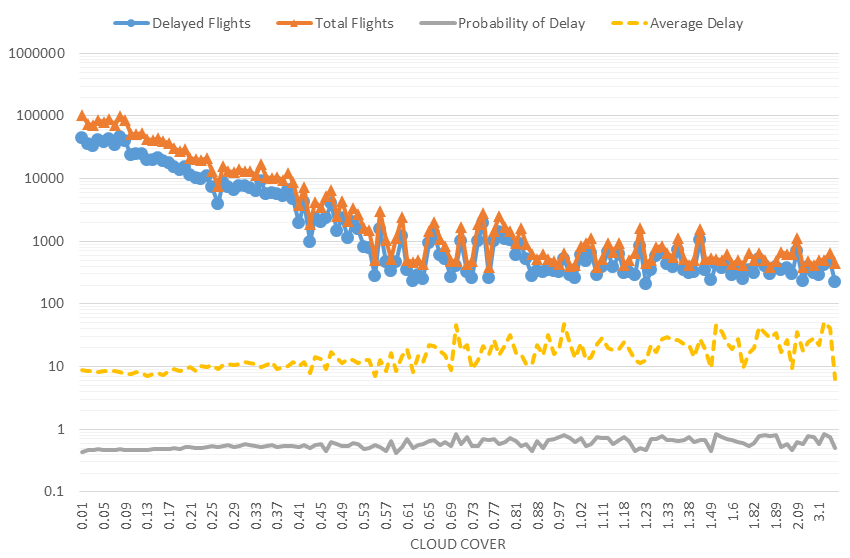}
\caption{Delay analysis on cloud cover. X-axis represents cloud cover and Y-axis represents number of delays.}
\label{9}
\end{figure}

\subsection{Joining the dataset}
MapReduce is used for this task. Our experiment is carried out by joining the above two described data sets to obtain a unified data set, which is used for the detailed analysis and prediction purpose. For this, two mappers and a single reducer are used. The first mapper takes input the airline delay data as input and the second mapper takes the weather data as input. The outputs of these mappers are then joined on the basis of the date, i.e. the key, and a new csv file is generated. This file is the unified one having all the required fields or attributes. It is this CSV file that is used for the whole experiment.

\subsection{Detail analysis}

For a manager of an airliner, there are an abundance of queries to be  performed regarding the delay. For instance, in which day, the probability of a flight delay is the highest? The result can be easily analyzed by the manager of the airline through the Figure ~\ref{1}. To get this result, we perform logical analysis on airline data and weather data generated as csv. The obtained csv file is used to analyze the delays in the airlines on the basis of different attributes to obtain the results shown in graphical form ~\ref{3},~\ref{4},~\ref{5},~\ref{6},~\ref{7},~\ref{8},~\ref{9}. We spawn many mappers and reducers to analyze the delay. Here, the mappers are given as input, the previously obtained CSV file in the KeyValueInputFormat. The mappers find out the delay for a particular key/value input pair, and output in the key/value format. The intermediate key is the value of the attribute under consideration and the value is the delay. The reducer fetches these key/value pairs in shuffled form and sums up all the delay values for a particular value of the attribute under consideration. The count of the total number of flights and the number of delayed flights is also maintained. Finally, the reducer outputs the attribute value, the total number of flights, the number of delayed flights as the key, the average delay and the probability of delay as the value. The Record Writer writes the final result in the mentioned output file in HDFS. 

The analysis results are plotted in graphical form as shown in Figure ~\ref{1},~\ref{2},~\ref{3},~\ref{4},~\ref{5},~\ref{6},~\ref{7},~\ref{8},~\ref{9}. Figures ~\ref{1},~\ref{2} do not consider the weather data. The figure ~\ref{1} reveals that the airline DL has the highest probability of delay  while average delay airline CO is the highest. Figure \ref{2} shows that most of the flight delays on Friday. However, other Figures ~\ref{3},~\ref{4},~\ref{5},~\ref{6},~\ref{7},~\ref{8},~\ref{9} combined airline data and weather data to analyze various scenarios. Figure ~\ref{3},~\ref{4},~\ref{5},~\ref{6},~\ref{7},~\ref{8},~\ref{9} are reported the delay analysis in the case of maximum temperature, minimum temperature, maximum visibility, minimum visibility, wind speed, precipitation, and cloud cover respectively. Thus, this analytics enhances the decision making process of the airline managers. Figure ~\ref{3} and ~\ref{4} reports the delay on maximum and minimum temperature in Fahrenheit. The maximum delay occurs in below 60 Fahrenheit in Figure ~\ref{3} and below 40 Fahrenheit in Figure ~\ref{4}. The figure ~\ref{5} and ~\ref{6} report the maximum delay in below 29.67 and below 2 respectively. Figure ~\ref{7} shows that the average delay become high when the mean wind speed high. Similarly, more cloud cover and precipitation cause delay as shown in Figure ~\ref{8} and ~\ref{9} respectively.

\section{Prediction}
\label{pr}

\begin{figure}[ht]
\centering
\includegraphics[width=0.5\textwidth]{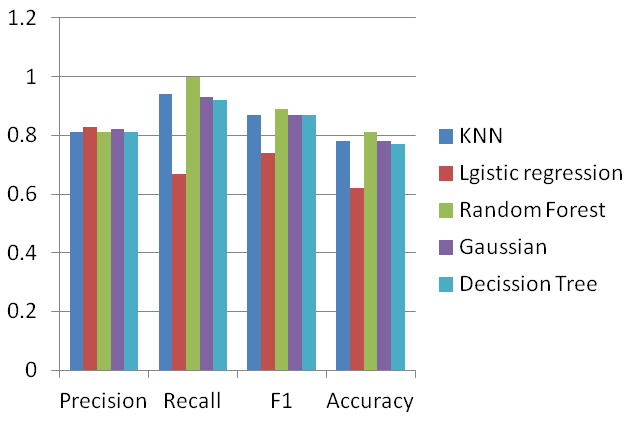}
\caption{Prediction analysis and results. X-axis represents precision, recall, F1 score, and accuracy and Y-axis represents score in an scale of 100.}
\label{pred}
\end{figure}

\begin{figure}[ht]
\centering
\includegraphics[width=0.5\textwidth]{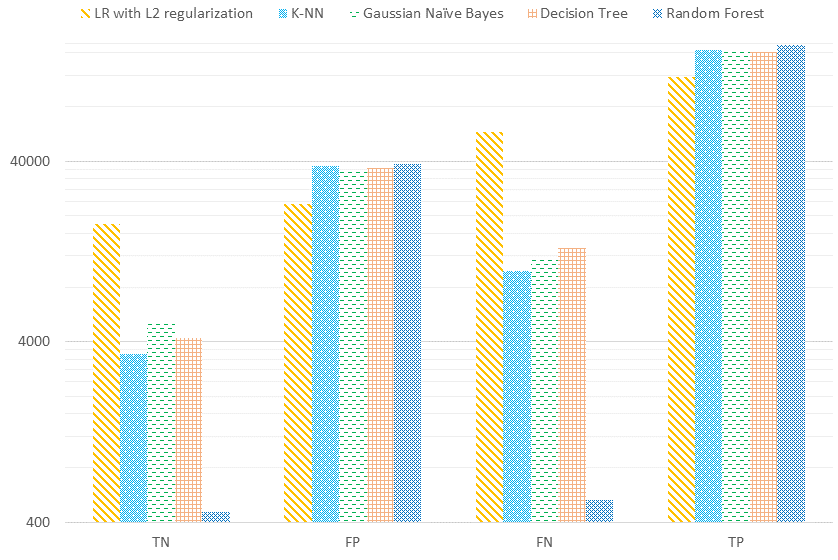}
\caption{Comparison among TN, FP, FN, and TN with machine learning model. X-axis represents TN, FP, FN and TN, and Y-axis represents counts on prediction.}
\label{cmp}
\end{figure}

For this part Python's NumPy and SciPy Library were used. Python's excellent Scikit-learn machine learning package is used to build the predictive models (Logistic regression, Random Forest, KNN, Gaussian Naive Bayes and Decision Tree Classifier) and compare their performance. First the previously mentioned CSV file (combination of weather and delay data) is loaded as feature matrix by filtering out all the non-numeric attributes. The matrix consists of all the data from the airline delay joined with the weather data. This feature matrix is of $(1566226) \times (8) $ size. The training dataset contained all the data from 2001 to 2007 and the test dataset contained the 2008 data. In the first Iteration Logistic Regression with L2 regularization was used. The delay is classified into binary classes:

\begin{itemize}
\item Delayed (more then 15 mins delay) 
\item Not Delayed (less then 15 mins delay).
\end{itemize}

After that, the Random forest model (50 trees) is used, it gives a better accuracy, and increase in true positives which signifies improvement in our model as shown in the table ~\ref{tab}. 

\subsection{Resultant confusion matrix}

\begin{figure*}[!ht]
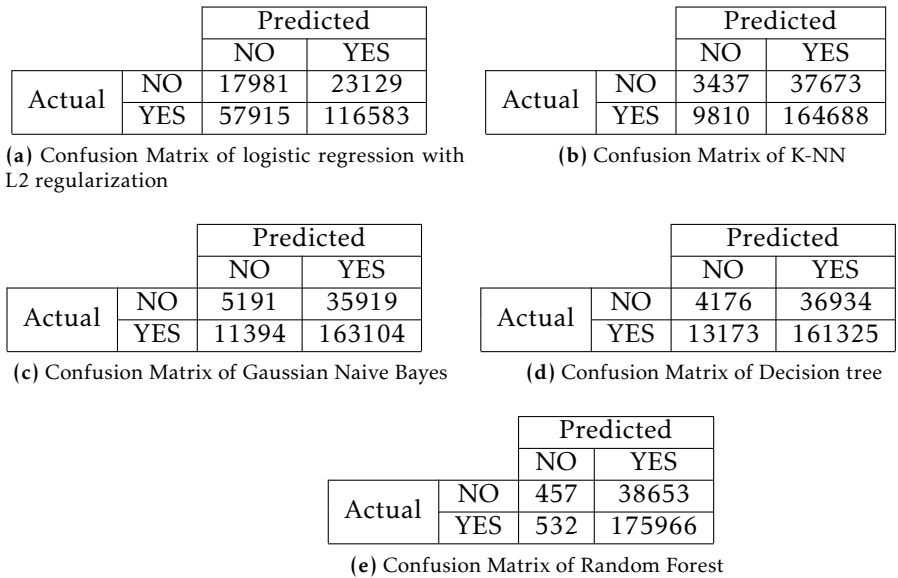

	\centering
	\subfloat[Confusion Matrix of logistic regression with L2 regularization\label{cm1}]{\taba}
    ~
    \subfloat[Confusion Matrix of K-NN\label{cm2}]{\tabb}
	\newline
	\subfloat[Confusion Matrix of Gaussian Naive Bayes\label{cm3}]{\tabc}
    ~
    \subfloat[Confusion Matrix of Decision tree\label{cm4}]{\tabd}
    \newline
    \subfloat[Confusion Matrix of Random Forest\label{cm5}]{\tabe}
    \caption{Confusion matrix}
\end{figure*}

The table ~\ref{cm1},~\ref{cm2}~\ref{cm3},~\ref{cm4} and ~\ref{cm5} show the confusion matrix after experimentation using logistic regression with L2 regularization, K-NN, Gaussian Naive Bayes, Decision tree and Random Forest model. The precision, recall, F1 score, and accuracy of logistic regression with L2 regularization are 0.83, 0.67, 0.74, and 0.62 respectively. The K-NN gives precision=0.81, recall =0.94, F1 score = 0.87, accuracy = 0.78. The Gaussian Naive Bayes gives precision = 0.82, recall = 0.93, F1 = 0.87, and accuracy = 0.78. The Decision tree gives precision = 0.81, recall = 0.92, F1 = 0.87, and accuracy = 0.77. The random forest achieves the highest and these are precision = 0.82, recall = 0.99, F1 = 0.90, and accuracy = 0.82.

Similarly, to these two models the other Machine Learning Techniques described above are also used for the same prediction purpose, and their results and performances are mentioned using tables and graphs. All these Machine Learning Techniques were used directly from the previously mentioned Python libraries. The total number of True Negative, False Positive, False Negative and True Positive is depicted in the figure ~\ref{cmp}.

\subsection{Prediction results and analysis}
We train using air traffic dataset from 2000 to 2007 and analyze the prediction using 2008 dataset. In this case, we omit all data from 1987 to 2000 to see how accurately can we predict airline delay with a very small set of data. We have split the training and testing dataset into 80:20 ratio. Surprisingly, the Random Forest model performs outstanding with an accuracy of 81\% using a very small set of data. The K-Nearest Neighbor and Gaussian Naive Bayes perform with an accuracy of 78\% while decision tree gives 77\%. We demonstrate the prediction results in figure ~\ref{pred}. As depicted in the figure ~\ref{pred},  we compare the performance of the Random Forest model, Logistic Regression With L2 Regularization, K-Nearest neighbor, Gaussian Naive Bayes model and Decision Tree model. In this comparison, the Random Forest model outperforms the other four models.

\section{Conclusion}
\label{con}
In this airline delay analysis and prediction, we have found that an airline can be scheduled in a better way if we uses analytics. As we have shown in the logical analysis reports, airline manager can perform analysis on various queries on delays. Moreover, the flight delay is predicted by training a very small set of data from 2000 to 2007 which is validated using 2008 dataset. Thus, a flight can be scheduled, organized and analyzed in a much better   way. The Random Forest outperforms other four machine learning models, however, K-NN and Gaussian Naive Bayes perform similar. The accuracy of the Random Forest model is 82\% with a threshold of 15 minutes. However, the work can also be carried out by Apache Mahout for machine learning library with Spark which is our future work. Finally, we believe that this work creates understanding in experimentation using weather and airline data for the development of new Airline Delay Analytics.

\bibliographystyle{abbrv}
\bibliography{mybibfile}

\end{document}